\documentclass{article}

\usepackage[dblblindworkshop,final]{neurips_2025} 
\workshoptitle{AI for Music}

\usepackage{authblk}
\usepackage[utf8]{inputenc} 
\usepackage[T1]{fontenc}    
\usepackage{hyperref}       
\usepackage{url}            
\usepackage{booktabs}       
\usepackage{amsfonts}       
\usepackage{nicefrac}       
\usepackage{microtype}      
\usepackage{xcolor}         
\usepackage{wrapfig} 
\usepackage{graphicx}
\usepackage{tcolorbox}
\usepackage{caption}
\title{MuCPT: Music-related Natural Language Model Continued Pretraining }

\author{%
  Kai Tian\textsuperscript{1}\thanks{This paper was completed by Kai Tian during his internship at Tencent.}\quad
  Yirong Mao\textsuperscript{2}\quad
  Wendong Bi\textsuperscript{2}\quad
  Hanjie Wang\textsuperscript{2}\quad
  Que Wenhui\textsuperscript{2}\thanks{Corresponding author.}\\
  \textsuperscript{1}\,Tsinghua University\\
  \textsuperscript{2}\,WeChat, Tencent Inc., Beijing, China\\
  \texttt{tk23@mails.tsinghua.edu.cn}\quad \texttt{\{erongmao,wendongbi,hankinwang,victorque\}@tencent.com}
}

\begin{document}

\maketitle

\begin{abstract}
Large language models perform strongly on general tasks but remain constrained in specialized settings such as music,particularly in the music–entertainment domain, where corpus scale, purity, and the match between data and training objectives are critical. We address this by constructing a large, music-related natural language corpus (40B tokens) that combines open source and in-house data, and by implementing a domain-first data pipeline: a lightweight classifier filters and weights in-domain text, followed by multi-stage cleaning, de-duplication, and privacy-preserving masking. We further integrate multi-source music text with associated metadata to form a broader, better-structured foundation of domain knowledge. On the training side, we introduce reference-model (RM)–based token-level soft scoring for quality control: a unified loss-ratio criterion is used both for data selection and for dynamic down-weighting during optimization, reducing noise gradients and amplifying task-aligned signals, thereby enabling more effective music-domain continued pretraining and alignment. To assess factuality, we design the MusicSimpleQA benchmark, which adopts short, single-answer prompts with automated agreement scoring. Beyond the benchmark design, we conduct systematic comparisons along the axes of data composition. Overall, this work advances both the right corpus and the right objective, offering a scalable data–training framework and a reusable evaluation tool for building domain LLMs in the music field.
\end{abstract}

\section{Introduction}

Large language models excel on broad text generation and understanding tasks, yet their effectiveness in specialized domains remains constrained by domain coverage and data quality\citep{dubey2024llama,jaech2024openai,team2024qwen2}. Music is a salient example: existing music-related natural language models demonstrate promise but are trained on relatively small or mixed-domain corpora, limiting factual coverage of artists, songs, and descriptors that matter in real applications\citep{yuan2024chatmusician}. To advance music-domain modeling, we  \emph{build a large music-related natural language dataset} and \emph{train on it with an objective aligned to the domain} so that specialization improves factual music QA without sacrificing general ability.

Our approach starts from data. We curate a \textbf{40B tokens} music-domain corpus including two parts: (1) \textbf{Matrix-music dataset} (20B tokens) and the \textbf{WeChat-music dataset} (20B tokens). The former is mined from Matrix\citep{zhang2024map} which is a public \emph{diverse and bilingual} pretraining dataset with roughly \emph{4.5T tokens}. The latter WeChat Dataset is curated from billions of WeChat articles and comment threads with entity-level links among singers, songs, and fine-grained passages. 

To effectively collect music-related data from the above huge text sources, we implement a scalable processing pipeline: a lite domain-first classifier is trained to filter and weight generic web data, a multi-stage cleaning stack removes duplicates and low-salience spans, and a high-relevance WeChat mining route with entity-level linking. The result is an end-to-end data foundation—the \textbf{Matrix-music dataset} plus the \textbf{WeChat-music Dataset}—that supports continued pretraining and instruction tuning in a unified and reproducible manner.

On the training side, we introduce a \emph{token-level soft scoring} method for fine-grained quality control, where a \emph{reference model} (RM) is first trained in a high-quality dataset and its per-token likelihoods are used to score the full corpus. In this way, we can (i) \emph{filter} low-quality tokens via a loss-ratio criterion and (ii) \emph{downweight} unreliable positions during optimization with an RM-normalized objective. 

To measure progress on factuality, we adopt MusicSimpleQA, a short-form, single-answer benchmark emphasizing verifiable facts\citep{wei2024measuring,li2023cmmlu,huang2023c,long2025adsqa,zeng2025reviewrl}. We use an automated agreement score to compute accuracy, enabling efficient and replicable evaluation. Empirically, our 32B domain-continued model (\textbf{Qwen2.5-32B-MuCPT}) reaches \textbf{0.7759} accuracy, surpassing GPT-4o\citep{hurst2024gpt}, Qwen3-235B-A22B-Instruct\citep{yang2025qwen3}, and DeepSeek-v3\citep{liu2024deepseek} on this task; notably, gains over the strongest baseline are +0.022 absolute (+2.92).

\begin{itemize}
    \item \textbf{A scalable music-domain data pipeline.} We combine large open and in-house sources with domain-first filtering, high-relevance mining, and entity-aware processing to construct a unified corpus for music continued pretraining. 
    \item \textbf{token-level soft scoring for quality control.} We propose an RM-based loss normalization and filtering strategy that removes or downweights low-quality tokens using a single scoring criterion for both selection and optimization. 
    \item \textbf{Factual music QA evaluation.} Using \textsc{MusicSimpleQA}, our 32B model achieves state-of-the-art accuracy among compared systems on factual music QA, and we provide training-paradigm and data-recipe comparisons that illuminate where gains come from. 
\end{itemize}

\section{Dataset Curation}
To enable continual pretraining in the music-related natural language domain, we adopt a domain-first data pipeline: a music-domain classifier is trained to filter and weight large generic corpora. We then apply multi-stage cleaning (language normalization, lightweight quality scoring, near-duplicate removal, privacy masking) to mitigate domain drift.

For the in-house WeChat Dataset, we anchor retrieval to top ten millions song names to mine related candidate articles, apply in-domain filtering, and perform entity-level alignment among singers, songs, and fine-grained passages. In parallel, we construct multi-source \emph{song–text} alignments: 3.5M song–snippet pairs covering 1.7M songs and comments coverage for 0.86M songs. Each song is further associated with weakly supervised tags (mood, genre, instrumentation, BPM) derived via a rules+statistics+LLM pipeline. We analyze the singer–song–document tri-graph and modestly upsample tails and new releases to curb head bias. This end-to-end pipeline provides fine-grained, verifiable supervision that matches the factuality emphasis of \textsc{MusicSimpleQA}. 
Full implementation details and statistics appear in the appendix~\ref{app:data-pipeline}.

\section{Token-level soft scoring and selection with Reference Model (RM)}
This section describes our token-level soft scoring for quality control for the pretraining corpus in the music domain. Building on the insight that not all tokens are equally useful, RHO-1\citep{lin2024rho} selects high-information tokens via a Reference Model (RM) and drop out uninformative tokens directly. We borrow this idea but go further by using RM scores to dynamically down-weight tokens in a soft way instead of filtering out tokens in a hard way in RHO-1. We follow a simple pipeline: (i) select a high-quality seed set, (ii) train a language model on this seed to serve as a RM that fits the target distribution, and (iii) use the RM to score tokens in the full corpus and apply filtering or weighting accordingly.

Formally, let $x_t$ denote the token at position $t$ and $x_{<t}$ its left context. We use the standard autoregressive negative log-likelihood (NLL) for the RM:
\[ \qquad
\mathrm{CE}_{\mathrm{RM}}(x_t) = -\log p_{\mathrm{RM}}(x_t \mid x_{<t}) .
\]
The RM is trained on high-quality seed set from WeChat wiki corpus of singers, songs and music-related entities. 

For the tokens from music-domain data, we apply RM-normalized loss that reduces the weight in a soft way when the RM loss is high. Specifically, the loss is
\[
L_{d}(x_t) \;=\; -\,\alpha \; \frac{\log p(x_t \mid x_{<t})}{\log p_{\mathrm{RM}}(x_t \mid x_{<t})} ,
\]
where $\alpha>0$ is a scaling coefficient. Intuitively, where the RM loss at a token is large, meaning that the token departs from the distribution of the high-quality seed set, the denominator increases and the effective contribution of that token is reduced. As the following simple example shows the gray noisy words would be down-weighted while training.
\begin{tcolorbox} 
\textbf{A real exemplar short article with noisy token:} <Try Everything> is a song performed by Colombian singer Shakira and serves as the theme song of the animated movie Zootopia. When I first watched Zootopia, I was deeply impressed by this theme song. After hearing the cover version by the One Voice Children's Choir, I realized that this song is not only catchy in melody, but also has lyrics that are youthful, sunny, and uplifting.
\textcolor{gray}{Click the border to bring up the video toolbar and scan the QR code to follow us to get more exciting content. Our address is: xxx, and phone is xxx.}

\label{tab:sample}
\end{tcolorbox}
\captionof{figure}{A real exemplar short article with noisy token}
To mitigate catastrophic forgetting in general knowledge, general-domain samples are mixed where we keep the usual autoregressive loss unchanged:
\[
L_{g}(x_t) \;=\; - \log p(x_t \mid x_{<t}) .
\]
Thus, the music-domain and general-domain samples are trained with different supervisions.

\section{Experiments}

\subsection{Factual Music QA on \textbf{MusicSimpleQA}: Setup and Findings}
\paragraph{Training Configuration.} Models are continual pre-trained in Qwen2.5 series models. Matrix-music dataset and WeChat-music dataset are combined as the final music-related natural language domain corpus which contains 40B tokens in total. To avoid catastrophic forgetting in general knowledge, data from general-domain pretraining dataset UltraFineWeb\citep{wang2025ultra} is sampled with meticulous mixture ratio tuning in small LLMs. We train for 2 epochs with 256 H20 GPUs, an initial learning rate of \(6\times 10^{-5}\), cosine scheduling with a minimum learning rate of \(3\times 10^{-5}\), and a warmup over the first \(0.05\%\) of steps.

\begin{wraptable}[9]{r}{0.55\textwidth}
\vspace{-0.8\baselineskip}  
\begin{tabular}{lc}
\hline
{\color[HTML]{333333} Models} & {\color[HTML]{333333} MusicSimpleQA} \\ \hline
GPT-4o                        & 0.6632                               \\
DeepSeek-v3                   & 0.7539                               \\
Qwen3-235B-A22B-Instruct      & 0.6719                               \\
Qwen2.5-32B-Instruct          & 0.3599                               \\ \hline
\textbf{Qwen2.5-32B-MuCPT}    & \textbf{0.7759}                     \\ \hline
\end{tabular}
\caption{Results on MusicSimpleQA (accuracy; higher is better).}
\label{tab:musicsimpleqa}
\end{wraptable}

\paragraph{Evaluation.}
We evaluate factual music knowledge with the \textbf{MusicSimpleQA} benchmark, which follows a short-form, single-answer design emphasizing verifiable facts. The final set contains 500 questions: 300 are drawn from currently popular artists to ensure practical relevance, and 200 are sampled by popularity evenly to broaden genre and era coverage. Evaluation is automated: a strong LLM (DeepSeek-v3) measures agreement between each model's prediction and the reference answer to produce an accuracy score. This construction enables efficient, replicable assessment of factuality in the music domain. 

\paragraph{Results.}

We compare GPT-4o, DeepSeek-v3, Qwen3-235B-A22B-Instruct, Qwen2.5-32B-Instruct\citep{qwen2.5}, and our domain-continued model \textbf{Qwen2.5-32B-MuCPT}. Results are shown in Table~\ref{tab:musicsimpleqa}: GPT-4o~0.6632, DeepSeek-v3~0.7539, Qwen3-235B-A22B-Instruct~0.6719, Qwen2.5-32B-Instruct~0.3599, and \textbf{Qwen2.5-32B-MuCPT}~\textbf{0.7759}. Relative to the strongest baseline (DeepSeek-v3), our model improves by +0.0220 absolute (+2.92\%). Gains vs.\ GPT-4o, Qwen3-235B-A22B-Instruct, and Qwen2.5-32B-Instruct are +16.99\%, +15.48\%, and +115.6\% respectively. These results indicate that domain-adaptive data and objectives can outweigh sheer parameter count on factual music QA.

Notably, a 32B model with strong domain adaptation outperforms a 235B instruction model and GPT-4o on this task, reinforcing the effectiveness of domain-incremental training when data and objectives are tightly matched to the target domain.

\subsection{Ablation study for our token-level soft scoring}

In this part, we compare our token-level soft scoring with plain next token prediction and RHO-1\cite{lin2024rho}. RHO-1 implements selective language modeling where a reference model scores tokens in the domain corpus with excess loss and drop out tokens in a hard way. The plain next token prediction treats every tokens equally.

\begin{wraptable}[12]{r}{0.55\textwidth}
\begin{tabular}{lc}
\hline
Method             & MusicSimpleQA   \\ \hline
Qwen-1.5B-NextTokenPrediction & 0.4439          \\
Qwen-1.5B-RHO-1    & 0.4759          \\
Qwen-1.5B-MuCPT (ours)    & \textbf{0.5259} \\
\hline
Qwen-7B-NextTokenPrediction   & 0.5499          \\
Qwen-7B-RHO-1      & 0.5699          \\
Qwen-7B-MuCPT (ours)      & \textbf{0.6119} \\ \hline
\end{tabular}
\caption{MusicSimpleQA accuracy for NextTokenPrediction, RHO-1, and MuCPT at 1.5B and 7B scales.}
\label{tab:rho1}
\end{wraptable}

Results are summarized in Table~\ref{tab:rho1}, under the same inference and evaluation protocol. At \textbf{1.5B} model scale, RHO-1 improves over \textit{NextTokenPrediction} (0.4439 $\rightarrow$ 0.4759; +7.21\%), while \textbf{MuCPT} reaches \textbf{0.5259} (+10.51\% vs.\ RHO-1; +18.47\% vs.\ NextTokenPrediction). At \textbf{7B} model scale, RHO-1 raises accuracy from 0.5499 to 0.5699 (+3.64\%); \textbf{MuCPT} attains \textbf{0.6119} (+7.37\% vs.\ RHO-1; +11.28\% vs.\ NextTokenPrediction).

Both RHO-1 and our token-level soft scoring in \textbf{MuCPT} surpass the next token prediction, showing that the not all tokens are equally contributed to the domain task. RHO-1 follows a ``select-before-learn'' principle that drops out noisy or uninformative tokens which may interrupting semantic coherence. Instead, our \textbf{MuCPT} retains token-level soft scoring for quality control where the semantic coherence is kept well.

\subsection{Comparing Data Recipes for Domain-Continued Pretraining}
We compare three different data sources in a fixed budget: 
(i) \textbf{BAAI/IndustryCorpus2-film-entertainment}\citep{beijing_academy_of_artificial_intelligence} (a film \& entertainment subset that is broader than music) , (ii) \textbf{Matrix-music dataset} \citep{zhang2024map}, and (iii) \textbf{WeChat-music dataset}. Results are reported in Table~\ref{tab:recipe-7b}.

\begin{wraptable}[8]{r}{0.45\textwidth}
\begin{tabular}{lc}
\hline
Models          & MusicSimpleQA \\ \hline
IndustryCorpus2\cite{beijing_academy_of_artificial_intelligence} & 0.2899        \\
Matrix-music (ours)     & 0.3659        \\
WeChat-music (ours)            & \textbf{0.4579}        \\ \hline
\end{tabular}
\caption{MusicSimpleQA accuracy with 7B-token data recipes for a 1.5B model.}
\label{tab:recipe-7b}
\end{wraptable}

\textbf{WeChat-music} achieves \textbf{0.4579}, whereas \textbf{Matrix-music} obtains 0.3659 and \textbf{IndustryCorpus2\_film\_entertainment} reaches 0.2899. Relative to Matrix-music, WeChat-music yields an absolute gain of 0.0920 (\,+25.14\%\,); relative to IndustryCorpus2\_film\_entertainment, the gain is 0.1680 (\,+57.95\%\,). Matrix-music also outperforms the broader film/entertainment subset by 0.0760 (\,+26.22\%\,).

Under the same token budget, the differences likely reflect a combination of factors such as task–distribution match, corpus quality and cleaning standards, sampling choices, and noise levels. In general, data recipes that align more closely with the knowledge distribution required by music QA tend to be more stable on \textsc{MusicSimpleQA}.

\section{Conclusion}

We presented a unified route to specialization in the music–entertainment domain that couples a \emph{domain-first} data pipeline with \emph{reference-model, token-level soft scoring for quality control}, and evaluates progress with the MusicSimpleQA benchmark for short-form factuality. The data pipeline prioritizes in-domain signals during sampling, integrates rigorous multi-stage cleaning and de-duplication, and organizes high-relevance sources into a coherent substrate for continued pretraining and alignment. The RM-normalized objective provides a single, per-token scoring principle that governs both selection and dynamic down-weighting, producing cleaner gradients and enabling efficient specialization while largely preserving general world knowledge. Taken together, these components deliver a scalable and auditable recipe for building music-domain LLMs that treat “music as a second language” without requiring task-specific modules.

\bibliographystyle{plain}
\bibliography{sample}


\appendix

\section{Related Work}
In recent years, propelled by breakthroughs in generative modeling, large language models (LLMs) have been introduced into the music domain, where incremental pre-training equips them with specialized musical knowledge and skills\citep{yuan2024chatmusician,ma2024foundation,zhang2025survey}. Continuing to train an LLM on domain-specific data—so-called domain-incremental training—has become the mainstream strategy for enhancing professional competence. For example, the ChatMusician\citep{yuan2024chatmusician} model is further pre-trained on top of LLaMA-2 and fine-tuned with the text-friendly ABC notation, enabling the model to treat music as a “second language” to be understood and generated. This approach exploits MusicPile\citep{yuan2024chatmusician}, a 4-billion-token music-language corpus that combines multimodal text—web-scraped music encyclopedias, music-theory books, YouTube metadata, and lyrics—with a large collection of scores encoded in ABC. To align musical and linguistic information, ChatMusician cleans and converts the data, using plain-text scores (ABC) to avoid extra multimodal modules. The model also adopts an incremental scheme that blends synthetic and public data, for instance augmenting the corpus with GPT-4-generated music Q\&A pairs and track summaries. A similar strategy appears in projects such as “MusicGPT”, which represent music as discrete symbol sequences and continue training a causal language model on those sequences: researchers feed large MIDI libraries, encoded with formats such as MMM Track, into a GPT-2-style model by flattening multitrack music into linear sequences\citep{ens2020mmm,pasquier2025midi,rhyu2024practical}. These results show that sustained pre-training on music data and careful data alignment allow an LLM to internalize music theory and symbolic structure, compressing and representing musical information without additional task-specific modules\citep{bhandari2025text2midi}.

Building on these pre-training strategies, LLMs have already demonstrated a range of capabilities in the music domain, spanning creation, analysis, and assistance. In music creation, large models can generate new compositions from user-provided text or other cues. MusicLM and MusicGen, for instance, enable text-to-music generation, producing audio in a requested style directly from natural-language descriptions\citep{copet2023simple,agostinelli2023musiclm}. Their outputs cover a wide spectrum—from Baroque choral works to modern pop—while maintaining musical correctness and coherence. LLMs similarly excel at lyric writing: by learning from vast lyric corpora, they can craft rhymed lyrics on a given topic or emotional tone\citep{ding2025songcomposer,liu2024agent,tian2023unsupervised}. In music analysis, LLMs demonstrate an understanding of and reasoning about musical knowledge—answering theory questions, parsing score structures, and recognizing style or affect after domain-incremental training. As auxiliary tools, large language models are becoming valuable assistants in music creation and education\citep{lei2024songcreator}. Conversationally, they retrieve musical information (e.g., historical facts, piece backgrounds) or act as virtual teachers answering theory queries; combined with tool invocation, they can call specialized audio-processing or generation models to accomplish complex tasks. Overall, whether composing autonomously or supporting human creativity and learning, LLMs are bringing the vision of “music as dialogue” to life.

\section{User Preference Analysis and the Motivation for Music–Entertainment CPT}
\label{app:user-preference}
\begin{figure*}[htbp]
    \centering
    \includegraphics[width=\linewidth]{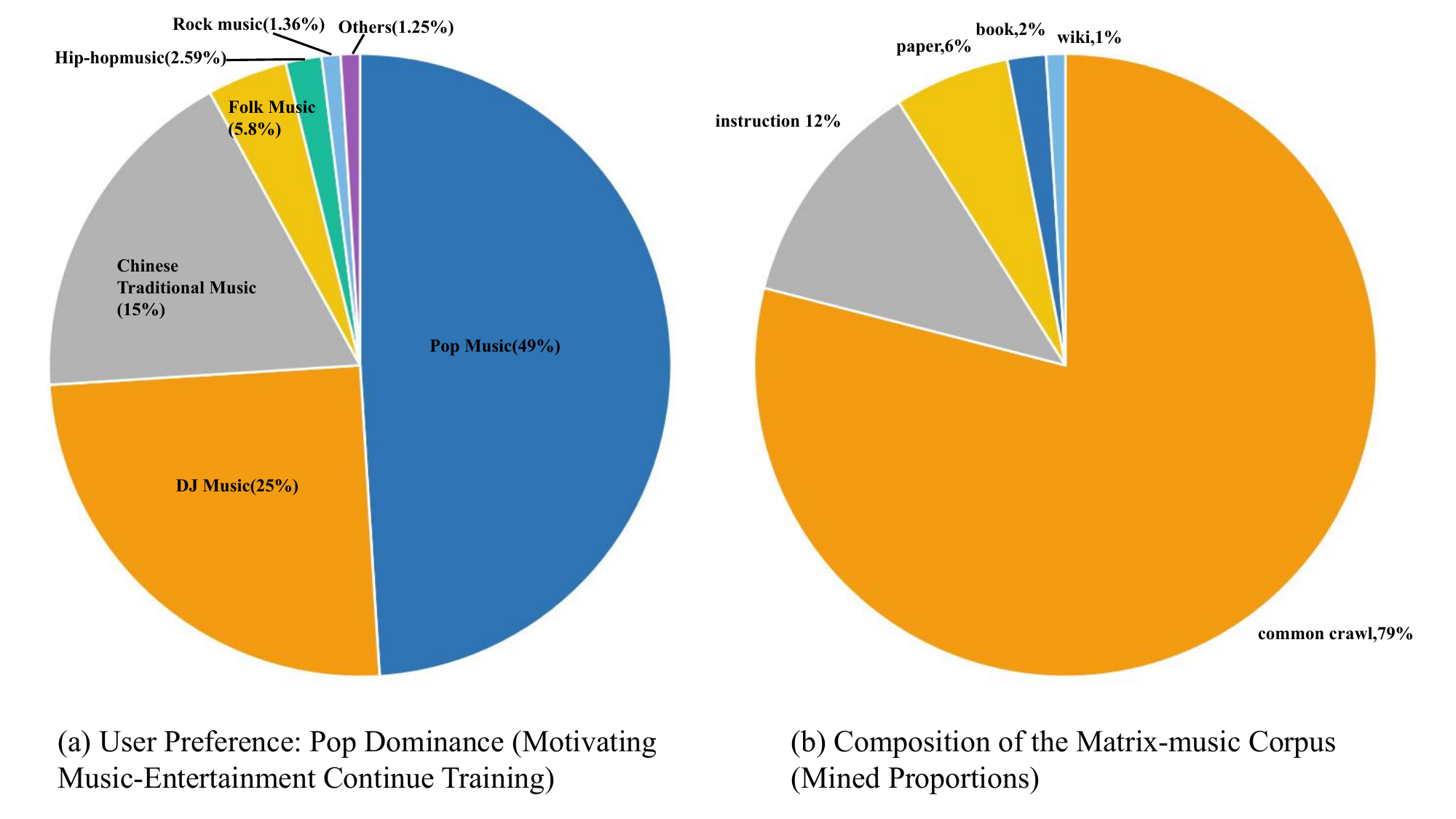}
    \caption{User preferences and corpus composition. (a) shows that most users prefer pop music, motivating our focus on music-entertainment continual pretraining; (b) summarizes the proportional makeup of the open-source music corpus mined for this work.}
    \label{fig:pie-user-pref}
\end{figure*}
Figure~\ref{fig:pie-user-pref} summarizes aggregate user-behavior signals (e.g., plays, searches, saves, skips) and shows that \textbf{pop music dominates overall preference}, far exceeding other categories (DJ/electronic, folk, hip-hop, rock, Chinese traditional, etc.). Guided by this observation, we prioritize \emph{music–entertainment} scenarios for continual pretraining (CPT) and construct our dataset accordingly: sources and samples that are closely tied to everyday entertainment-oriented music consumption receive higher sampling priority and stricter cleaning. 

\section{Details of Data Curation}
\label{app:data-pipeline}
Our objective in this stage is continual learning for the music domain. We adopt a domain-first strategy: we first train a domain classifier and use it to filter and weight large-scale generic corpora before continued pretraining and task-oriented augmentation. The classifier is based on Qwen2.5-0.5B and trained as a balanced binary model with roughly 250k positive and 250k negative examples. Positive instances include content about songs, artists, catalogs, reviews, styles, and production notes; negatives span diverse non-music topics. During sampling, the classifier’s confidence serves as a routing signal that assigns higher weights to in-domain text and gates expensive instruction-style or alignment-oriented augmentations.

For the Matrix-music dataset, we aggregate and clean about 20 billion tokens of open data .  Book contributes roughly 332 million tokens, common crawl about 17.4 billion, instruction about 2.7 billion, paper about 1.3 billion, and wiki about 270 million. Cleaning follows a multi-stage pipeline: language identification and normalization; heuristic and lightweight quality scoring to prune templated or low-salience spans; locality-sensitive hashing/MinHash for near-duplicate removal; and privacy-preserving masking for potential identifiers. 

Figure~\ref{fig:pie-user-pref} presents the source composition of the cleaned and normalized Matrix-music corpus: Common Crawl dominates (79\%), followed by instruction (12\%), paper (6\%), book (2\%), and wiki (1\%). The figure characterizes the \emph{base distribution} adopted after a multi-stage cleaning pipeline (“broad coverage + lightweight quality scoring + near-duplicate removal + privacy masking”). 

To strengthen high-relevance Chinese coverage, we build a WeChat-music Dataset via a mining pipeline. Using approximately ten millions top song names from the last year (from a wechat-listen source) as anchors, we retrieve and match candidate articles, apply the music classifier for in-domain filtering, and then perform entity-level alignment that links singers and song titles to fine-grained passages with disambiguation. This pipeline yields roughly 20 billion tokens across about 20.5 million documents. We analyze the singer–song–document tri-graph to understand coverage and long-tail effects, and we modestly upsample tail and newly released works in subsequent sampling to reduce head-content dominance.

To directly learn alignments between text and musical works, we construct song–text pairs from multiple sources. On the comment side, we select reliable public comments and filter coarse noise and toxicity with an LLM, resulting in coverage for roughly 0.86 million songs; these naturally encode perceptual descriptors such as timbre, mood, style, and instrumentation, providing supervised attribute-alignment signals. On the article side, we segment articles into fragments, yielding about 3.5 million “song–snippet” pairs that cover approximately 1.7 million songs. For each song, we further generate or aggregate content-understanding tags—mood, genre, instrument, and BPM—using a multimodal music understanding model; tagging follows a hybrid “rules + statistics + LLM cross-check” procedure: lexicon- and rule-based retrieval to form candidates, statistical co-occurrence and contrastive models for scoring, and an LLM pass for consistency and readability checks.

In summary, we link preference-driven domain filtering, scalable high-relevance mining, entity-level alignment with verifiable task construction. The $\sim$22B-token base corpus ensures breadth, the $\sim$21B-token WeChat-music Dataset supplies high-relevance Chinese context, and multi-source song/comment/article pairs plus mood/genre/instrument/bpm tags provide fine-grained supervision.

\section{Construction of the Music Factuality Question-Answer Evaluation Set}

In the construction of the music factuality question-answer evaluation set, we drew inspiration from OpenAI’s SimpleQA evaluation set and the approach outlined in \cite{wei2024measuring}, with the aim of developing a tool capable of efficiently and automatically evaluating the performance of large language models in the music domain — the MusicSimpleQA evaluation set. The core design of this evaluation set centers around the concept of "factuality," constructing a series of concise, clear, and unique question-answer pairs, ensuring that each question has a single, unambiguous answer, which is crucial for assessing the model’s ability to handle real-world factual knowledge. 

To generate the questions, we first employed the Deepseek-v3 model to automatically extract information from extensive singer encyclopedic data and generate common factual questions related to music artists. These questions cover basic biographical and factual information about the artists, such as "Where is Jay Chou from?", "What is the name of Jay Chou's first solo album?", "Which film did Jay Chou win the Best Newcomer Award at the Golden Horse Awards?", and "Which talent show did Zhou Shen debut in?". The answers to these questions are  unique and verifiable facts.

Following the generation of questions, we conducted a filtering and validation process to ensure that the answers were both unique and clear. Specifically, all generated questions underwent manual or automated consistency checks, where those questions that had multiple possible answers or were ambiguously phrased were filtered out, ensuring that only questions with clear and unique answers remained. This filtering process ensures that the evaluation set maintains high quality and precision, preventing ambiguous or non-unique answers from interfering with the evaluation results.

The final evaluation set consists of 500 questions. Of these, 300 questions are based on popular artists within the current music industry to ensure that the evaluation set is relevant and realistic. The remaining 200 questions are popularity evenly sampled, covering different genres and eras, to increase the diversity of the set and to assess the model’s ability to handle a wide range of musical knowledge. The evaluation method involves comparing the standard answer with the model’s predicted answer. The DeepSeek-v3 model is used to automatically evaluate the consistency between the standard and predicted answers. Specifically, DeepSeek-v3 compares the matching degree between the given standard answers and the model-generated answers, providing an automatic score for the prediction’s accuracy.

By employing this automated and efficient evaluation method, we provide a new evaluation tool for large language models in the music domain — MusicSimpleQA. This evaluation set allows us to comprehensively assess the performance of models in answering factual questions in the music domain. The construction of this evaluation set not only contributes to the development of models in the music field but also provides a replicable framework for future music data processing and knowledge inference tasks.

\section{General-Ability Evaluation: Impact on World Knowledge}

\begin{table}[htbp]
\centering
\begin{tabular}{lll}
\hline
Models               & C-Eval & CMMLU \\ \hline
Qwen2.5-32B-Instruct & 86.46  & 85.79 \\
Qwen2.5-32B-MuCPT    & 84.17  & 84.34 \\ \hline
\end{tabular}
\caption{World-knowledge accuracy on \textsc{C-Eval} and \textsc{CMMLU} for Qwen2.5-32B-Instruct vs.\ Qwen2.5-32B-MuCPT.}
\label{tab:world-knowledge}
\end{table}

We assess whether music-domain continual pretraining harms broad world knowledge by comparing \textbf{Qwen2.5-32B-Instruct} and \textbf{Qwen2.5-32B-MuCPT} on two comprehensive benchmarks, \textsc{C-Eval} and \textsc{CMMLU}. As summarized in Table~\ref{tab:world-knowledge}, MuCPT shows a small drop on \textsc{C-Eval} (86.46 $\rightarrow$ 84.17; absolute $-2.29$, relative $-2.65\%$) and a similarly minor drop on \textsc{CMMLU} (85.79 $\rightarrow$ 84.34; absolute $-1.45$, relative $-1.69\%$). Overall, both changes are within $<3\%$, indicating no catastrophic forgetting of world knowledge. In practice, MuCPT maintains general ability while delivering substantial gains on music-domain factual QA, demonstrating a favorable specialization–generalization trade-off.


\end{document}